\documentclass[conference,10pt]{IEEEtran}
\usepackage{amssymb,amsmath,color}
\usepackage{graphicx}
\pdfoutput=1
\usepackage{breqn}
\usepackage[export]{adjustbox}
\input{epsf}
\usepackage{subfigure}
\usepackage{fancyvrb}
\usepackage{alltt}
\usepackage{balance}
\usepackage{graphicx}

\usepackage{algorithm}
\usepackage{algorithmic}
\usepackage{hyphenat}
\usepackage{multirow}

\usepackage[center]{caption}
\usepackage{url}
\usepackage{hyperref}
\usepackage{wrapfig,lipsum,booktabs}
\DefineVerbatimEnvironment{code}{Verbatim}{fontsize=\small}
\usepackage{cite}
\graphicspath{{./eps/}}
\DeclareGraphicsExtensions{.eps}
\usepackage{mdwmath}
\usepackage{mdwtab}
\usepackage{eqparbox}
\usepackage{wrapfig,lipsum,booktabs}
\usepackage{scrextend}

\begin{document}
%
%================================================================%
%							                TITLE
%================================================================%
%
\title
  {
    Co-evolving Real-Time Strategy Game Micro
  }

%================================================================%
%							                AUTHORS
%================================================================%
%
%
\author
  {
	  \IEEEauthorblockN
    {
    Navin K Adhikari\IEEEauthorrefmark{1},
    Sushil J. Louis\IEEEauthorrefmark{2}  
    Siming Liu\IEEEauthorrefmark{3}, and
    Walker Spurgeon\IEEEauthorrefmark{4}
    }
    \IEEEauthorblockA
    {
    Department of Computer Science and Engineering                      \\
    University of Nevada, Reno                                          \\
    Email:                                        %
    \IEEEauthorrefmark{1}navinadhikari@nevada.unr.edu,  %
    \IEEEauthorrefmark{2}sushil@cse.unr.edu
    \IEEEauthorrefmark{3}simingl@unr.edu, and
    \IEEEauthorrefmark{4}wspurgeon@nevada.unr.edu
    }
  }
\maketitle
%
%================================================================%
%                            ABSTRACT
%================================================================%
%
\begin{abstract}

We investigate competitive co-evolution of unit micromanagement in
real-time strategy games. Although good long-term macro-strategy and
good short-term unit micromanagement both impact real-time strategy
games performance, this paper focuses on generating quality
micro. Better micro, for example, can help players win skirmishes and
battles even when outnumbered. Prior work has shown that we can evolve
micro to beat a given opponent. We remove the need for a good opponent
to evolve against by using competitive co-evolution to evolve
high-quality micro for both sides from scratch. We first co-evolve
micro to control a group of ranged units versus a group of melee
units. We then move to co-evolve micro for a group of ranged and melee
units versus a group of ranged and melee units. Results show that
competitive co-evolution produces good quality micro and when combined
with the well-known techniques of fitness sharing, shared sampling,
and a hall of fame takes less time to produce better quality micro
than simple co-evolution. We believe these results indicate the
viability of co-evolutionary approaches for generating good unit
micro-management.

\end{abstract}
\vspace{-5pt}
%  
%
%================================================================%
%                            INTRODUCTION
%================================================================%
%
%%%%%%%%%%%%%%%%%%%%%%%%%%%%%%%%%%%%%%%%%%%%%%%%%%%%%%%%%%%%%%%%%%%%%%%%%
\section{Introduction}
\label{SectionIntroduction}
%%%%%%%%%%%%%%%%%%%%%%%%%%%%%%%%%%%%%%%%%%%%%%%%%%%%%%%%%%%%%%%%%%%%%%%%%

Real-Time Strategy (RTS) games have become a new research frontier in
the field of Artificial Intelligence (AI) as they represent a
challenging environment for an autonomous agent. An RTS game player
needs to collect resources, use those resources to construct a base,
train units, research technologies and control different types of
units to defeat the opponent while at the same time defending their
own base from opponent attacks in a complex dynamic environment. All
of these different actions that can be executed in any given state
make a huge decision space for a player. RTS players usually divide
these decision spaces into two different levels of tasks:
macromanagement and micromanagement. Macromanagement encompasses a
wide variety of tasks such as collecting more resources, constant unit
production, technology upgrades, and scouting. In contrast,
micromanagement is the ability of a player to control a group of units
to beat an opponent. Better micro, for example, can help players win
skirmishes and battles even when outnumbered or minimize damage
received. Although good long-term macro-strategy and good short-term
unit micromanagement both impact real-time strategy games performance,
this paper focuses on generating quality micro.  More specifically, we
focus on two aspects of the micromanagement: tactics and reactive
control~\cite{ASORSGARACIS13}. Tactics deal with the overall
positioning and movement of a group of units while reactive control
deals with controlling a specific unit to achieve commonly used micro
techniques: concentrating fire on a target, retreating seriously
damaged units from the front line of the battle, and kiting (hit and
run).

We build on prior work ~\cite{EEMIRSG16} and represent micro-behaviors
of units on both sides with a set of parameters. We use a commonly
used technique called Influence Maps (IMs) to represent enemy
distribution over the game map. An IM is a grid placed over the map
with a value assigned to each grid cell using an IM function that
depends on the number of enemy units in the vicinity. Good IMs can
tell us where the enemy is strongest and where they are weakest (That
is, the best target position for friendly units to go). To navigate a
group of units to a target location on the map given by an IM, we use
Potential Fields (PFs). PFs are used widely in multi-agent systems for
coordinated movement of multiple
agents~\cite{CCNMAS07}~\cite{FCMC01}. We use two IMs parameters and
four PF parameters for tactics and six parameters for reactive
control. We then use a Co-evolutionary Genetic Algorithm (CGA) to
search and find good combinations of these twelve parameters that lead
to a good micro behavior on both sides.

Prior work has shown a Genetic Algorithm (GA) can evolve good micro to
beat a given opponent but that micro performance depends on having a
good opponent to play against. Furthermore, it is non-trivial to
hard-code a good opponent to play against. We remove the need for a
good opponent to evolve against by using competitive co-evolution to
evolve high-quality micro for both sides from scratch. In competitive
co-evolution, two populations evolve against each other. That is,
individuals in one population play against individuals in the other
population for evaluating each other's fitness. Note that the fitness
of an individual in one population depends on the fitness of the
opponents drawn from the other population. As both population evolve,
individuals from each population must compete against more and more
challenging opponents leading to an “ arms race”. This simple model of
co-evolution suffers from several well known
problems~\cite{schonfeld32survery}. Although even this simple model of
co-evolution works well enough to produce better than random micro, we
use three techniques: competitive fitness sharing, shared sampling and
hall of fame from Rosin and Belew~\cite{NMFCC97} to produce better
quality micro in less time than using simple co-evolution.
 
We first co-evolve micro to control a group of ranged units versus a
group of melee units. We then move to co-evolve micro for a group of
ranged and melee units versus an opponent group of ranged and melee
units. Results show that we can co-evolve good micro for both
opponents in both scenarios. In addition, we tested generalizability
of the co-evolved micro by evaluating performance of co-evolved micro
in different initial configurations and different initial positions of
units. Results show that micro co-evolved in one scenario work well in
other scenarios as well.

%% One of the major challenges in competitive co-evolution is the
%% large computational burden. This is because every individual needs
%% to compete against individuals from the other population to measure
%% fitness. In our simulation engine, it takes an average of five
%% seconds for one evaluation. Although, shared sampling decreases
%% computational effort, we were able to co-evolve we decreased the
%% computation effort using hall of fame and shared sample, evaluation
%% time is still large. We therefore ran CGA in parallel on multiple
%% cores to make the evaluation process faster.

%% This is not research
%% and since everyone who uses GAs runs in parallel, there is nothing
%% new or noteworthy in running in parallel

The remainder of this paper is organized as follows. Section
~\ref{SectionRelatedWork} describes related work in RTS AI research,
generating game players using co-evolutionary techniques and common
techniques used in RTS micro. The next section describes our RTS
research environment. Section ~\ref{SectionMethodology} explains our CGA
implementation. Section ~\ref{SectionResultsDiscussion} presents preliminary
results. Finally, the last section provides conclusions and discusses
possible directions for future work.
%
%================================================================%
%                            RELATED WORK
%================================================================%
%%%%%%%%%%%%%%%%%%%%%%%%%%%%%%%%%%%%%%%%%%%%%%%%%%%%%%%%%%%%%%%%%%%%%%%%%
\section{Related Work}
\label{SectionRelatedWork}
%%%%%%%%%%%%%%%%%%%%%%%%%%%%%%%%%%%%%%%%%%%%%%%%%%%%%%%%%%%%%%%%%%%%%%%%%

Traditionally, much work has been done in Computational Intelligence
(CI) and AI in games revolving around board games, using a variety of
techniques~\cite{AMLIGAS01}~\cite{ACPAR11}. More recently, research
has shifted away from board game towards more complex computer games
and real-time strategy games like starcraft pose several challenges
for computational intelligence research~\cite{RSANARC03}. In addition, 
challenges in RTS games are strikingly similar to real-world
challenges making RTS games a good research platform.

Much work has been done on RTS games addressing different aspects of
developing an AI player for such games ~\cite{ASORSGARACIS13}. In this
paper, we are interested in one aspect of an RTS game: ``Micro.'' Micro
stands for micromanagement, the reactive and tactical control of a
group of units to maximize their effectiveness (usually) in combat.
Game tree search techniques and machine learning approach have been explored for
micro tactics and reactive control. Churchill and Buro explored a game tree
search algorithm for tactical battles in RTS
games~\cite{FHSFRTSGCSS12}. Synnaeve and Bessiere applied bayesian
modeling to inverse fusion of the sensory inputs of the units for
integration of tactical goals directly in
micro-management~\cite{ABMFRUCATS11}. Wender and Watson evaluated
different reinforcement learning for
micro-magement~\cite{ARLTSSCITRTSGSB12}.

%%%%%%%%%%%%% Don't use et al

A number of micro techniques use influence maps and potential fields
for tactical and reactive control of units in an RTS game. An
Influence Map (IM) tiles a map into square tiles with each tile or
grid cell getting a value provided by an IM function. Grid cell values
determine enemy unit locations and concentrations and can be used to
provide a variety of useful information for unit maneuvering. We
describe influence maps and potential fields later in this
paper. Miles evolved the parameters of an influence map using a
genetic algorithm in order to evolve an RTS game
player~\cite{CIMTBSGP07}. Sweetser and Wiles used IMs to help a
decision-making agent in their EmerGEnt
game~\cite{CIMACAFRGA05}. Bergsma and Spronck generated adaptive AI
for a turn-based strategy game using IMs ~\cite{ASRFTSG08}. Jan and
Cho used information provided by layered IMs to evolve nonplayer
characters' strategies in the strategy game
Conqueror~\cite{ENNWLIMITRSGC08}. Preuss investigated flocking-based
and IM-based path-finding algorithms to optimize group movement in the
RTS game Glest ~\cite{CATITCAMIRSG10}~\cite{IMOGIRSG08}. Uriarte and
Ontanon used an IM-based approach to evolve kiting (similar to
hit-and-run) behavior in the Starcraft bot
Nova~\cite{KIRGUIM12}. Danielsiek investigated influence maps to
support flanking the opponent in a RTS game~\cite{IMOGIRTSG08}. This
paper uses influence maps to determine a target location to move
towards and attack. Potential fields guide our unit movement.

Potential fields (PFs) of the form $cd^e$ where $d$ is distance and
$c$ and $e$ are tunable parameters have been used in robotics and
games for generating smooth group movement~\cite{ROAFMAMR86}. Relevent to
RTS games, Jonas and Kostler used PFs to control units optimally in
StarCraft II for simulating optimal
fights~\cite{IAGAFSOFIS16}. Sabdberg and Togelius Hagel investigated
multi agent potential field based AI approach for small scale combat
in RTS games~\cite{EMPFBAAFSSIRG11}. Rathe and Svendsen did unit
micromanagement in Starcraft using potential
fields~\cite{MISUPFTWMOGA12}. They all used genetic algorithm to tune
multiple potential fields' parameters. Hagelback and Johansson applied
potential fields in their RTS games research~\cite{UMPFIRSG08}. They
proposed a multiagent PF-based bot architecture for the RTS games ORTS
and applied PFs for tactical and reactive unit movement. 
Closer to our work, Liu and Louis used parameterized algorithms that
determined unit positioning, movement, target selection, kiting, and
fleeing. Then a genetic algorithm tuned these parameters by evolving
against a hand-coded opponent or an existing Starcraft BWAPI
bot~\cite{EEMIRSG16}. We build on this prior work and use the same
representation (parameterized algorithms) but co-evolve, rather then
evolve, micro without the need for a good opponent to play
against.

Coevolution in games goes back to Shannon's work on checkers in the
50s with the most recent notable example being Alpha-go
Zero~\cite{GPM95}~\cite{silver2016mastering}. In RTS games,
Ballinger and Louis showed that coevolution led to more robust build
orders. Build-order optimization enables players to generate the right
mix and numbers of units meeting a strategic
need~\cite{CCGAAHFFRSGP13}. Avery and Louis coevolved team-tactics
using a set of IMs, navigating a group of friendly units to move and
attack enemy units on the basis of the opponent's
position~\cite{CIMFSTTIATG10}. More relevant to our coevolutionary
approach, Rosin and Belew improved the performance of a coevolution
using three techniques: competitive fitness sharing, shared sampling
and hall of fame~\cite{NMFCC97}. We use these techniques and show
their effectiveness in coevolving good micro in less time than simple
coevolution.

The next section describes our game engine and provides details on how
we simulate skirmishes in this game engine for fitness evaluation. We
then describe our representation and evolutionary algorithm tuned
parameters and our methodology for measuring coevolutionary progress.

%======================================================================================%
%                            Methodology
%======================================================================================%
%%%%%%%%%%%%%%%%%%%%%%%%%%%%%%%%%%%%%%%%%%%%%%%%%%%%%%%%%%%%%%%%%%%%%%%%%
\section{Methodology}
\label{SectionMethodology}
%%%%%%%%%%%%%%%%%%%%%%%%%%%%%%%%%%%%%%%%%%%%%%%%%%%%%%%%%%%%%%%%%%%%%%%%%

\begin{figure}%[htp]
\centerline{
  \includegraphics[width=3.5in]{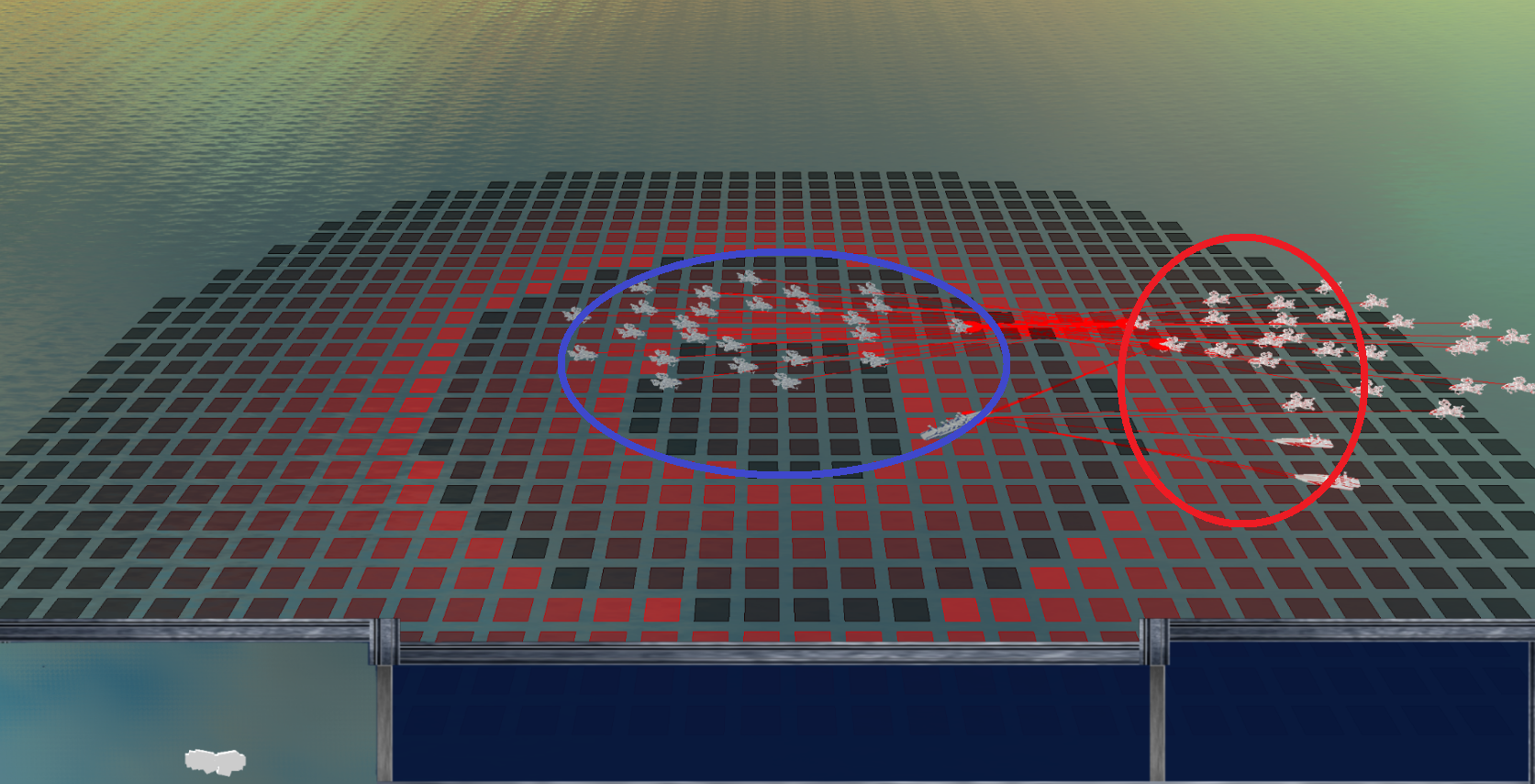}
  }
\caption{Screenshot of a skirmish in FastEcslent}
\label{Figure1}
\end{figure}

Apart from the Starcraft BWAPI and Starcraft II API, there now exist a
number of other RTS game-like engines that can be used for RTS game
research~\cite{heinermann2012bwapi}~\cite{website2014}. The open-source FastEcslent game engine
which runs game graphics in a separate thread is especially suitable
for evolutionary computing research in games since we can run the
underlying game simulation without graphics and thus more easily do
multiple parallel evaluations. Figure~\ref{Figure1} shows a screenshot
from FastEcslsent running with graphics.

We use unit health, weapons, and speed values from
Starcraft to create the equivalent of Vultures and Zealots in
FastEcslent~\cite{website2013}~\cite{heinermann2012bwapi}.  A Vulture is StarCraft unit
that is fast but fragile and can attack form a longer distance which
helps such units ``kite,'' during a skirmish while a Zealot is slower
but stronger and has a shorter attack distance. To evaluate the
fitness of a chromosome, we decode the chromosome and use the twelve
resulting parameters to control our units in the game simulation. The
simulation ends when all the units on one side are destroyed or time
runs out. After each simulation, FastEcslent returns a score for each
side in the skirmish based on how much damage was done and how much
damage was received and this score is used to compute a fitness.

\begin{figure}
\centerline{
 \includegraphics[width=3.5in]{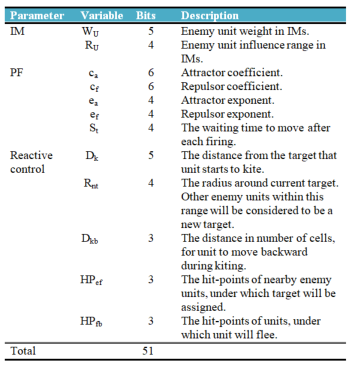}
}
\caption{Parameters tuned by coevolution}
\label{TwelveParameters}
\end{figure}

The goal of our work is to evolve good micro for opponents in an RTS
game without the need of an opponent to evolve against. We therefore
use a coevolutionary algorithm to achieve this goal. In coevolution,
two populations of individuals play each other to compute fitnesses
that drive evolution~\cite{NMFCC97}. Extending prior work, we
represent micro by a set of parameterized algorithms and the genetic or
coevolutionary algorithm tunes these parameters to find good
micro. These algorithms specify group positioning, movement, target
selection, kiting, and fleeing.  Table~\ref{TwelveParameters} details
the twelve parameters in our representation which is identical to the
representation used by Liu~\cite {EEMIRSG16}. Tuning these parameters
results in micro for one type of friendly unit against one type of
enemy unit. We explain these parameters below.

%\begin{table}[H]
%\caption{Chromosome representing each individuals for Population I and Population II}
%\end{table}

Good positioning during a skirmish can reduce damage received and
increase damage dealt to opponent units. We use Influence Maps (IMs)
to try and find vulnerable positions to attack. An influence map is a
grid of cells placed over the map, where each cell has a value
determined by an IM function. In our work, the IM function specifies a
weight parameter ($W_e$) for each cell occupied by an enemy
entity. The entity’s influence decreases as a function of distance and
ceases after $R_e$ distance (in number of cells) units. We sum the influence of all enemy
entities in range of cell to compute the cell’s IM value. The cell
with the lowest value and that is closest to the enemy determines our
attack location. The GA or coevolutionary algorithm determines $W_e$
and $R_e$.

How a group of units moves to the target location also determines
skirmish performance. We use attractive and repulsive potential fields
to control group movement~\cite{}. The typical representation of an
attractive and a repulsive potential field is given by equation~\ref{PFEquation}
\begin{equation}
\label{PFEquation}
\mathcal PF = c_ad^{e_a}+c_rd^{e_r}
\end{equation} 
where $C_a$ and $E_a$ are parameters of the attractive force and $C_r$
and $E_r$ are parameters of repulsive force. However, balancing these
forces to achieve smooth, effective unit movement is difficult and we
therefore use the CGA to find the best values for these
parameters. Once we reach the target location, target selection,
kiting, and fleeing become important. Good target selection can
significantly affect performance since selecting a weaker target, to
destroy more quickly, can thus more quickly reduce damage being
received. The CGA evolves the two parameters, $HP_ef$ and $R_nt$,
defined in Table~\ref{TwelveParameters} to guide a target selection
algorithm~\cite{EEMIRSG16}.

Kiting by longer ranged units is an effective tactic used by good
human players in skirmishes with short ranged melee units. Three
parameters that determine kiting behavior are 1) how far away from a
target the unit needs to start kiting ($D_k$), 2) the waiting time
before moving after each firing ($s_t$), and 3) how far a unit should
retreat before attacking ($D_kb$). We use a parameterized kiting
algorithm which uses these three parameters to
kite~\cite{EEMIRSG16}. Finally, removing weakened units from the front
line to save them for later is determined by a hit-point threshold
$HP_fb$, also coevolved by the CGA. Good values for all these
parameters can lead to micro that beats state of the art BWAPI
competition bot micro~\cite{RSANARC03}~\cite{EEMIRSG16} when evolved
against such micro. This paper seeks to use CGAs to reach high levels
of performance without the need for good micro to evolve against.

%======================================================================================%
%                               Fitness Evaluation     
%======================================================================================%
\subsection{Coevolution and Fitness Evaluation}
\label{SubsectionFitness Evaluation}
In coevolution, individual fitnesses result from direct competition
between individuals in two populations. We want to maximize damage
done and minimize damage received. More precisely, when an individual
$i$, from one population competes against individuals from the other
population, $i$ get’s a score given by equation~\ref{score}, based on damage
done by $N_f$ friendly units to $N_e$ enemy units and damage received in each competition.
\begin{equation}
  \label{score}
  \begin{array}{lcl}
    \mathcal Score & = & V_1 \sum\limits_{n=1}^{N_f} (HP_f/HPF_{max})\\
                   & + & V_2 \sum\limits_{n=1}^{N_e} (HPE_{max} - HP_e)
    \end{array}
\end{equation}
$HPF_{max}$ is the starting hitpoints corresponding to maximum health
for each friendly unit. Similary $HPE_{max}$ specifies the starting
hitpoints of each enemy unit. $HP_f$ represents the remaining
hitpoints for friendly units at the end of a fitness simulation while
$HP_e$ represents the same parameter for enemy units. $V_1$ and $V_2$
are scores for saving friendly hitpoints (health) or reducing enemy
hitpoints.  We obtain these values from the Starcraft BWAPI. We
explain how this score leads to an individual's fitness after
describing the coevolutionary algorithm.

\begin{figure}
\centerline{
  \includegraphics[width=3.5in]{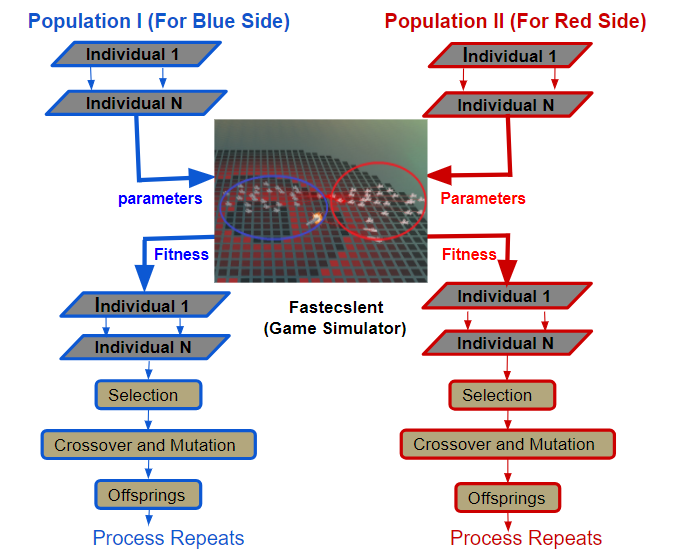}
  }
\caption{The coevolutionary algorithm plays individuals from the blue
  population against individuals in the red population to obtain
  damage done and received and thus determine relative fitness.}
{\label{CGA}}
\end{figure}

Since we are coevolving both sides, we refer to the two sides
coevolving in their distinct populations as red and
blue. Figure~\ref{CGA} shows how individuals in the blue and red
populations are evaluated and how a single evaluation determines the
fitness of two individuals - one from the blue and one from the red
population.

\subsection{One unit type versus one unit type}

Our first experimental scenario coevolved $5$ red Vultures against
$25$ blue Zealots. For each individual in the blue population, we send
the $12$ parameters specified by that individual to control micro for
the $25$ blue side zealots against every individual in the red
population. Each red individual's chromosome controls the $5$
vultures. For a population size, $p$, and assuming both red and blue
have the same population size, we need a total of $p^2$ evaluations to
obtain a fitness for every individual in both
populations.

Equation~\ref{score} specified the score received during one
evaluation; an individual's fitness is the average of all the scores
obtained by playing one red individual (for example) against all $p$
members of the blue population. $V_1$ and $V_2$ differ for the red and
blue populations since each is trying to micro a different type of
unit.  $V_1 = 400, V_2 = 160, HPF_{max} = 80, HPE_{max} = 160, N_e =
25$ (zealots), and $N_f = 5$ vultures for the red population. From the
blue population's point of view, these parameter values are
different. Blue friend Zealots compete against red enemy Vultures and
$V_1 = 160, V_2 = 80, HPF_{max} = 160, HPE_{max} = 80, N_e = 5$
(vultures), and $N_f = 25$ zealots for the blue population. Except for
the number of enemies, $N_e$, and number of friends, $N_f$, the values
of all other parameters are obtained from the Starcraft1.

\subsection{Two unit types versus two unit types}

Good results from coevolving micro for groups composed from one type
of unit versus groups also composed from one, albeit differnt, type of
unit led us to consider a second set of experiments where we
investigated coevolving micro for groups composed from two types of
units against an opponent group also composed from two types of
units. Specifically, we coevolved micro for a group of $5$ vultures
and $25$, say on the red side against an identical group of $5$
vultures and $25$ zealots on the blue side. Our chromosomes doubled in
size from $12$ to $24$ parameters and the first $12$ parameters
controlled vultures while the second set of $12$ parameters controlled
zealots.

We also generalized Equation~\ref{score} to handle multiple types of
friend and enemy units. Essentially this means that there are two
values for $V_1$, one for vultures ($400$) and one for zealots
($160$). Similarly there are two values for $V_2$ when considering
damage to enemy vultures ($80$) and zealots ($160$). Maximum values
for hitpoints also depend on the unit type.

For simple competitive coevolution, the fitness of an individual is
the average of scores obtained from “playing” against all individuals
in the opponent population. An individual “plays” against another by
being placed in our game and running the game until either all the
units in one side are destroyed or time runs out. Once we have such a
measure of fitness, the two populations can potentially coevolve
leading to an “arms-race” of increasing fitness.

Although this model of coevolution works well enough to produce better
than random micro, we use three techniques: competitive
fitness sharing, shared sampling, and hall of fame as described by
Rosin and Belew~\cite{NMFCC97} to produce better quality micro in less
time than using simple coevolution~\cite{hillis1990co}. We provide
brief descriptions of these three methods below.

The idea of fitness sharing is to prevent diverse niches from
prematurely going extinct. Sharing an individual’s score from
defeating a specific individual $i$ drawn from the opponent population
among all the individuals that defeated $i$, leads to higher fitness
for individuals that defeat opponent individuals that no one else
can. This decreases the probability of important innovations going
extinct.
 
The usual way to evaluate an individual is to play against all the
individuals in the opponent population. To reduce computational
effort, shared sampling evaluates an individual by playing against a
sample of individuals drawn from the opponent population. In order to
increase the diversity in this opponent sample, first select an
opponent individual ‘A’ that defeated the most individuals in your
population. Then an individual defeating those individuals that
defeated ‘A’ are selected, and so on, until the sample size becomes
full.

A finite population means that a high fitness individual from one
generation may not stay high fitness in a different context provided
by an evolving opponent population. To ensure against permanent loss
of such strong individuals and to prevent cycling caused by
intransitive superiority, we keep such current strong
individuals in a hall of fame so that we can use a strong diverse
sample of past (hall of fame) and current individuals to play against
in order to gain a better measure of an individual’s fitness. This
helps evolve individuals that are more robust.

%======================================================================================%
%                                         RESULTS AND DISCUSSION
%======================================================================================%
%%%%%%%%%%%%%%%%%%%%%%%%%%%%%%%%%%%%%%%%%%%%%%%%%%%%%%%%%%%%%%%%%%%%%%%%%
\section{Results and Discussion}
\label{SectionResultsDiscussion}
%%%%%%%%%%%%%%%%%%%%%%%%%%%%%%%%%%%%%%%%%%%%%%%%%%%%%%%%%%%%%%%%%%%%%%%%%
%

For all evaluations, we ran for a maximum of $2500$ frames which,
despite running without graphics, took an average of $5$ seconds per
evaluation. We therefore parallelized evaluations to get reasonable
run times and achieved approximately linear speedup. Coevolution run results are
averaged over ten runs with different random seed.

First, we coevolved micro for a group of ranged units versus a group
of melee units with simple coevolution - that is coevolution without
any shared sampling, shared fitness, or hall of fame.  Second, we
compared the results produced using simple coevolution with the
results produced when using all three techniques. 

Specifically, in the first set of experiments we coevolve $5$ red
vultures versus $25$ blue zealots. Both populations used a population
size of $50$ and we ran for $60$ generations. Crossover and mutation
probabilities were $0.95$ and $0.03$ respectively.

Since the fitness of an individual, $i$, depends on the quality of
individuals from the opponent population that $i$ competes against in
coevolution, plotting fitness over time does not measure progress in
the same that such plots do for standard genetic algorithms. Instead,
we use a different approach and start by generating a baseline
individual.  Every coevolutionary generation, we take the best
individual from the blue (or red) population and play this best
individual against the fixed baseline. As coevolution progresses, we
expect the best individual in subsequent generations to improve
performance over the fixed baseline.

\begin{figure}
    \centerline{
      \includegraphics[width=3.5in]{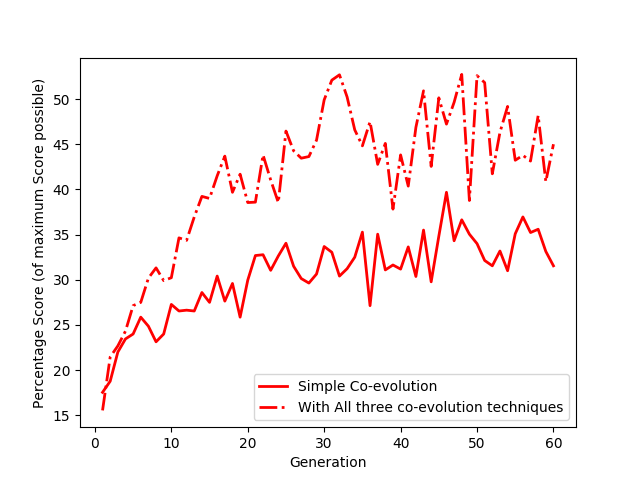}
      }
    \caption{Performance of coevolving vultures against the baseline zealots}
    \label{CoVulturesVsZealots}
\end{figure}

Figure~\ref{CoVulturesVsZealots} plots the best coevolving vulture
(red population) against such a baseline zealot. This baseline zealot
beats $1996/2000$ randomly generated individuals for a $90\%$ win rate.
The solid line shows simple coevolution while the dashed line shows
coevolution augmented with fitness sharing, shared sampling, and hall
of fame. We can see improvement over time for both and we can see that
the three techniques do improve micro quality faster.  To reduce the
computational effort, the size of shared sample and hall of fame
should be as low as possible. But, decreasing the size too much may
reduce needed diversity in the set of individuals selected for playing
against. We thus need a delicate balance between maintaining diversity
in the shared sample and hall of fame to play against, and low
computational effort. In these and subsequent experiments the shared
sample size and hall of fame size are both set to five ($5$) - a value
found through experimentation. With these settings we get $50$
population size $\times 10$, shared sampling size plus hall of fame
size, $\times 2$ for the two populations for a total of $1000$
evaluations per coevolutionary generation. This equates to a savings
of \( \frac{2500 - 1000}{2500} = 60\% \) in terms of computational
effort measured in number of evaluations.

\begin{figure}
    \centerline{
      \includegraphics[width=3.5in]{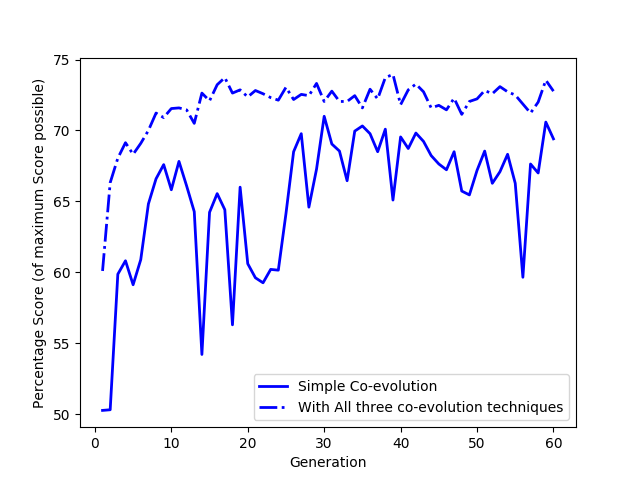}
      }
    \caption{Performance of coevolving zealots against the baseline vultures}
    \label{CoZealotsVsVultures}
\end{figure}

Figure~\ref{CoZealotsVsVultures} shows a similar patterns when
comparing the coevolving zealots against a baseline vulture micro.
Note that in both sets of results using the three methods result in
smoother performance curves. Videos of gameplay show complex patterns
of movement. Zealots learn to herd vultures into a corner while
vultures learn kiting and to stay out of range of zealots. These
videos are available at
\href{https://www.cse.unr.edu/~navin/coevolution}{https://www.cse.unr.edu/~navin/coevolution}.

Building on these results, we next investigated more complex micro for
groups composed from zealots and vultures versus an opponent also
built from the same two types of units. With two types of units we can
also look to see whether, and what kind of, cooperative behavior
emerges between the two types of units.

\subsection{Two types of units versus two types of units}

We investigate coevolving micro for a group of $5$ vultures and $25$
zealots versus an identical opponent group ($5$ zealots, $25$
vultures). We did not test simple coevolution, preferring to use the
coevolution with the three methods since the augmented coevolutionary
algorithm performs better and due to a lack of time.

Again, we used a population size of $50$ running for $60$ generations
with the same crossover and mutation probabilities of $0.95$ and
$0.03$ as before. Note that the chromosome size needs to double so
that micro for the two unit types coevolve to take advantage of each
unit type's unique properties. Progress is again measured against a
baseline group of $5$ vultures and $25$ zealots that beats $99\%$ of
randomly generated opponents.

\begin{figure}
    \centerline{
      \includegraphics[width=3.5in]{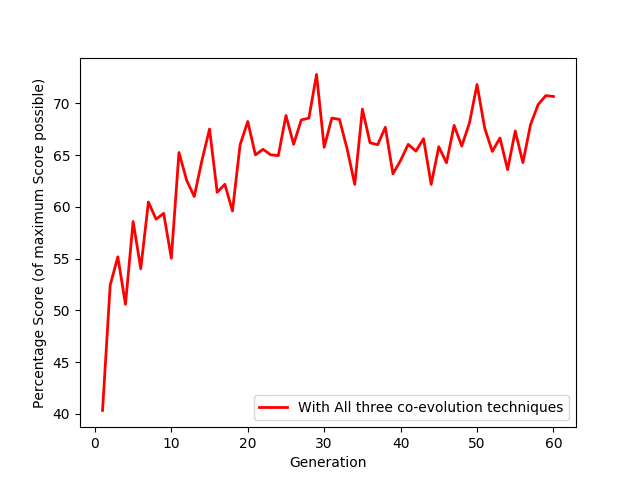}
      }
    \caption{Performance of coevolving zealots and vultures versus the baseline for the Red population}
    \label{CoZealotsVulturesPopRed}
\end{figure}
\begin{figure}
    \centerline{
      \includegraphics[width=3.5in]{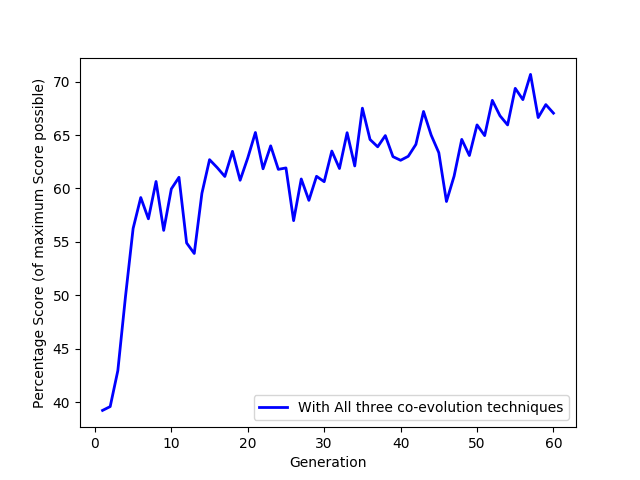}
      }
    \caption{Performance of coevolving zealots and vultures versus the baseline for the Blue population}
    \label{CoZealotsVulturesPopBlue}
\end{figure}

Figure~\ref{CoZealotsVulturesPopRed} shows coevolutionary progress
against this baseline for the red population. Again we see fairly
smooth (for coevolution) progress in finding increasingly good
micro. Figure~\ref{CoZealotsVulturesPopBlue} shows, unsurprisingly, that the coevolving blue
population has similar performance improvement.
These results seem to indicate the potential for a coevolutionary
approach to coevolve good micro from scratch.

%%%%%%%%%%%%%%%%%%%%%%%%%%%%%%%%%%%%%%%%%%%%%%%%%%%%%%%%%%%%%%%%%%%%%%%%%%%%%%%%%%%%%%%%
%%%%%%%%%%%%%%%%%%%%%   Sushil %%%%%%%%%%%%%%%%%%%%%%%%%%%%%%%%%%%%%%%%%%%%%%%%%%%%%%%%%

%% Third, we used our approach to coevolved
%% micro-strategy for mix type units on both sides; that is a group of
%% ranged and melee units versus a group of melee units.  We then checked
%% the generalizations of our coevolved micro-strategy into different
%% initial group arrangements, different initial positions and different
%% group combinations.
%%%%%%new newe results%%%%%%%%%

Next we consider the robustness of this coevolved micro by testing the
coevolved micro in scenarios hitherto unseen. First, we looked at the
micro coevolved for the one unit type versus one unit type experiments
and selected the best co-evolved individual from both populations. We
then played these two best individuals in three different starting
formations (or scenarios) and ten different starting locations. We did the same for
the best individuals in the two unit types versus two unit types
experiments.

\begin{figure}
    \centerline{ \includegraphics[width=3.5in]{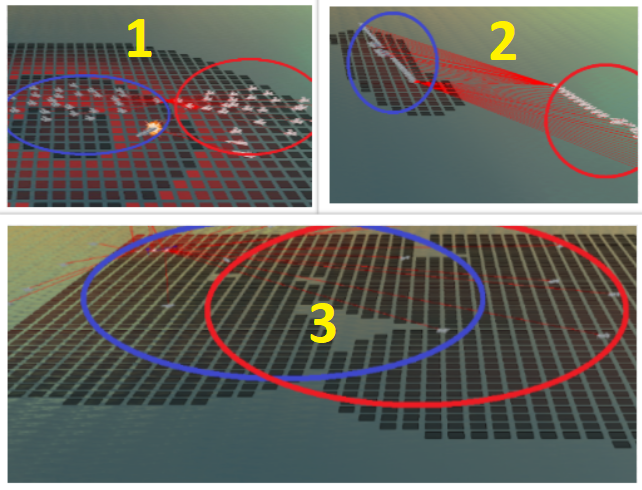} }
    \caption{Snapshot of circular formation (1), line formation (2), and random formation (3)}
    \label{formations}
\end{figure}

Figure~\ref{formations} shows screenshots of these three scenarios. The first distributes units within a circle (labeled 1), the second uses a line formation (2), and the third distributes units randomly (3). Blue and Red indicate side in the screenshot.
We describe our experiments and results with respect to these three
formations next. Red bars represent red population (vulture)
performance versus baseline zealots and blue bars represent blue
(zealot) performance versus baseline zealots.

\subsection{Scenario 1: Circular}
\begin{figure}
    \centerline{
      \includegraphics[width=3.5in]{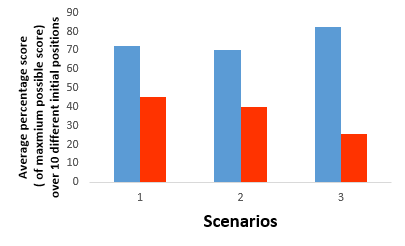}
      }
    \caption{Performance of co-evolved player against baseline in different scenarios; Co-evolved Vulture (Red), Co-evolved Zealot (Blue)}
    \label{barOneVOne}
\end{figure}

This was our training scenario in that coevolution took place with
units placed within this circle and always started in the same initial
positions during a fitness evaluation. To test robustness we randomly
changed the starting positions and generated $10$ randomly generated
sets of starting positions for the units. We tested the coevolved
micro against our baseline player on these 10 different scenarios and
computed the average score. Figure~\ref{barOneVOne} shows that
coevolved micro seems robust to starting position with performance
similar to those in Figure~\ref{CoVulturesVsZealots} and
Figure~\ref{CoZealotsVsVultures}.

\subsection{Scenario 2: Line formation}
In this scenario, units from both sides area placed in a line opposite
each other on the game map. With this formation, we want to see how
the coevolved micro does when changing both the formation and the
initial positions on this formation. Again, we randomly generated 10
different sets of unit starting positions on the line and averaged the
score obtained by the best coevolved micro against our baseline.  The
second set of bars in Figure~\ref{barOneVOne} shows that co-evolved
micro does just as well in this new formation over multiple sets of
starting locations.

\subsection{Scenario 3: Random starting locations}
In this scenario, rather than putting units into any particular
formation, we place them randomly in the game.  The score of the best
individual against baseline player is again averaged over 10 different
initial position and shown in Figure~\ref{barOneVOne}. With this
formation, we can see that vultures do not fare well. We address this
in our future work.

\begin{figure}
    \centerline{
      \includegraphics[width=3.5in]{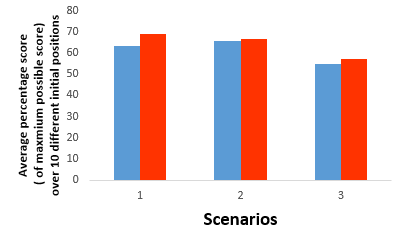}
      }
    \caption{Performance of co-evolved player against baseline in different scenarios; Red and Blue sides}
    \label{barTwoVTwo}
\end{figure}

Finally, Figure~\ref{barTwoVTwo} shows the same information for the two unit types versus two unit types experiments. 
We can see the same trend. These figures indicate the potential for our CGA approach to find good robust micro from scratch. 
%
%======================================================================================%
%                                                                                      CONCLUSION
%======================================================================================%
%%%%%%%%%%%%%%%%%%%%%%%%%%%%%%%%%%%%%%%%%%%%%%%%%%%%%%%%%%%%%%%%%%%%%%%%%
\section{Conclusions and Future Work}
\label{SectionConclusion}
%%%%%%%%%%%%%%%%%%%%%%%%%%%%%%%%%%%%%%%%%%%%%%%%%%%%%%%%%%%%%%%%%%%%%%%%%

Our research focuses on exploring coevolutionary approaches to finding
good micro in RTS games. This eliminates the need for a good opponent
to evolve against. We compactly represented micro with 12 parameters
that control simple algorithms for target selection, kiting, and unit
movement and used a coevolutionary algorithm to tunes these parameters
values. We measured the performance of two independent coevolving
populations by playing the best individual from each generation and
each population against a baseline player seperately. Results show
that we can coevolve a group of ranged units versus a group of melee
units using simple coevolution. We also compare these results using
three different techniques for improving competitive coevolution as
described by Rosin and Belew~\cite{NMFCC97}. Results also indicates
that we can coevolve a better micro in less time than using simple
coevolution.

We then coevolved micro for units composed from two types of units
versus similar opponents. For a mix of ranged and melee units results show that we can
coevolve good micro-behavior using coevolution augmented with shared sampling, fall of fame, and shared fitness. 

Both sets of solutions seem to be robust. We checked the robustness of
our co-evolved micro in three different unseen scenarios and ten
different sets of starting positions. Results shows that our approach
can find micro that performs well in unseen scenarios. We believe,
using a combination of random and structured scenarios during
coevolution will lead to more robustness.
 
The main constraints with a coevolution in an RTS game is
computational effort for evaluations. As a single fight simulation
takes significant time and each individual needs multiple evaluations,
required computational effort tends to outstrip available
resources. Although shared sampling and hall of fame result in good
reduction, it still takes days to get significant results. Given more
computational resources, we may be able to use much larger population
sizes and much longer run times to get significantly higher quality.

We believe these results indicate the viability of coevolutionary
approaches for generating good unit micromanagement and we plan to
build on this in our future work. We would like to investigate other
representations and coevolve within Starcraft II using the recently
released Starcraft II API.

%================================================================%
%                        REFERENCES
%================================================================%
%
%{\scriptsize
\bibliographystyle{IEEEtran}
\balance
\bibliography{IEEEabrv,cig2018}
%}
%
\end{document}